\documentclass[a4paper]{scrartcl}

\usepackage{times}
\usepackage{helvet}
\usepackage[english]{babel}
\usepackage[utf8]{inputenc}

\usepackage{authblk}
\usepackage[nolist, nohyperlinks]{acronym}
\usepackage{natbib}
\usepackage[binary-units=true]{siunitx}
\usepackage{amsmath}
\usepackage{amssymb}
\usepackage{latexsym}
\usepackage{graphicx}

\usepackage[vlined, ruled, commentsnumbered, linesnumbered]{algorithm2e}

\usepackage[hidelinks]{hyperref}
\usepackage{cleveref}

\DeclareSIUnit\au{arb. unit}

\begin{document}

\title{Active exploration of sensor networks from a robotics perspective}

\author{Christian Blum}
\author{Verena V. Hafner}
\affil{Adaptive Systems Group\\ Department of Computer Science\\ Humboldt-Universit\"at zu Berlin, Germany\\ email: blum@informatik.hu-berlin.de}

\date{}

\maketitle

\begin{abstract}
Traditional algorithms for robots who need to integrate into a wireless network
often focus on one specific task. In this work we want to develop simple,
adaptive and reusable algorithms for real world applications for this scenario.
Starting with the most basic task for mobile wireless network nodes, finding
the position of another node, we introduce an algorithm able to solve this
task.  We then show how this algorithm can readily be employed to solve a large
number of other related tasks like finding the optimal position to bridge two
static network nodes. For this we first introduce a meta-algorithm inspired by
autonomous robot learning strategies and the concept of internal models which
yields a class of source seeking algorithms for mobile nodes. The effectiveness
of this algorithm is demonstrated in real world experiments using a physical
mobile robot and standard 802.11 wireless LAN in an office environment. We also
discuss the differences to conventional algorithms and give the robotics
perspective on this class of algorithms.  Then we proceed to show how more
complex tasks, which might be encountered by mobile nodes, can be encoded in
the same framework and how the introduced algorithm can solve them.  These
tasks can be direct (cross layer) optimization tasks or can also encode more
complex tasks like bridging two network nodes. We choose the bridging scenario
as an example, implemented on a real physical robot, and show how the robot can
solve it in a real world experiment.
\end{abstract}

\section{Introduction}

Autonomous mobile network nodes are becoming more and more commonplace in the
form of robots. Robots turn into mobile network nodes when they need to
communicate data with either other robots, infrastructure or humans. For
example robot swarms often rely on explicit communication between each other
and base stations and can even span their own network
\citep{daniel2010communication, hauert2010communication} or they want to make
use of the sensors of a sensor network or even everyday objects like lighting
systems or even a fridge, which become increasingly smarter and networked
\citep{mattern2010internet}. This also includes semi-autonomous operation of
robots like for example in disaster scenarios in a supporting role for
firefighters or rescue crews.

This means that they have to deal with the issues of signal attenuation, noise
and interference, which are all of a spatial nature. A mobile robot can thus
also use its mobility to mitigate negative effects in a cross layer fashion.
This can in principle encompass everything from bandwidth optimization over
energy savings up to interference mitigation.

We show one possible way to make use of the robot's mobility to optimize one
network parameter and go on to show how this approach can be extended to all
kinds of other criteria or parameters. We do this explicitly from a
robotics point of view in order to give some new insights into this
cross-cutting issue between networking and robotics.

\subsection{State of the Art}

When dealing with any kind of taxis algorithm, the first question is if (local)
position information is available to the agent or not. If there is no position
information available, the only possible class of algorithms are stochastic
ones similar to the ones employed by nature for example in chemotaxis of
bacteria \citep{berg1972chemotaxis}. When using directional antennas, this
limitation can be mitigated and simple Braitenberg-style algorithms
\citep{braitenberg1986vehicles} can find a source. However in this paper we are
more interested in algorithms using position information but only measuring
scalar samples of the target function\footnote{This allows us to extend the
target function to more complex targets, see \cref{sec:extensions}.}.

Many algorithms estimate the bearing to sources from measurements by
estimating \ac{RSSI} gradients \citep{dantu2009relative, han2009access} or
exploiting the anisotropic radiation profiles of antennas
\citep{derenick2011localization}.  Furthermore, \ac{RSSI} gradients can be used
for frontier exploration of coverage areas \citep{twigg2012rss}. These gradients
can be estimated either using classical finite difference methods or by for
example fitting a plane to local measurements to mitigate noise and using
this plane to estimate the gradients \citep{paul2011radio}.

Gradient based methods can be proven to converge for the case of signal
strength in wireless networks under some constraints
\citep{atanasov2012stochastic,blum2014gradient}. These methods basically
implement gradient descent directly on noisy measurements and need to mitigate
local maxima created by small scale fading.

More complex models than local linear approximations have been shown to
effectively estimate source localization \citep{fink2010online}. This algorithm
uses Gaussian process models based on path-loss and attenuation priors. It
performs very well in terms of accuracy and statistical efficiency, i.e., the
number of samples needed but it unfortunately scales cubic with the number of
samples.

Also more complex algorithms for source seeking have been presented. For
example Wadhwa et al. \citep{wadhwa2011following} present a multi-phase
heuristic algorithm to mitigate the effects of very flat gradients far away
from the source. This algorithm has only been evaluated in simulation.

\section{Algorithm}\label{sec:algorithm}

We first present a high-level view of the (meta-) algorithm we designed. This
algorithm was designed with the task of finding the source of a wireless signal
in mind but can readily be extended to a host of different tasks as will be
shown in \cref{sec:extensions}.

The algorithm is presented schematically on a high level of abstraction in
\cref{alg:main}. While executing the algorithm, the robot receives a constant
and asynchronous stream of messages from its various sensors, of which the
current 2D position and the signal strengths of measured packets are used by
the algorithm. At the beginning of each iteration of the algorithm, these
messages are parsed and the positions interpolated to fit the timestamps of the
signal strength measurements. The robot is capable of localizing and navigating
autonomously which also means that movements are restricted to legal movements,
i.e., ones which are not part of a wall or outside of the known map.  For
details refer to \cref{sec:hardware_ros}.

\begin{algorithm}[h]
\LinesNotNumbered 
\KwIn{x,y,r as asynchronous messages}
\KwOut{move towards source}

do random movement\;
\While{not at source}
{
  parse messages\;
  \If{abs(mean RSSI over last second - model prediction at current position) $>$ error threshold}
  {
    discard current model\;
    learn new model on all available data\;
  }
  \eIf{random() $>$ $\epsilon$}
  {
    go with min(dist to goal, step width) towards global minimum of the model\;
  }
  {
    do random movement\;
  }
}
\caption{Main Algorithm}
\label{alg:main}
\end{algorithm}

The robot starts the algorithm with a random movement, which corresponds to a
flat prior on all possible options. The initial movement is restricted to twice
the regular step width and facilitates a first learning of the internal signal strength
model (see \cref{sec:internal_models}). We chose a constant step width of
$1\,m$ for all experiments.

At the beginning of each iteration, all messages in the buffer are parsed and
added to the available dataset. Then the model prediction error at the current
location is calculated against the mean signal strength over the last second to
mitigate some of the noise. If the prediction error surpasses an error
threshold (we used a threshold of 3dB), the current model is discarded and a
new model is learnt based on all available data. In principle the model could
be discarded in every iteration and replaced with a new model based on the latest
data but in order to keep computational cost to a minimum, especially for more
complex models, we stick to a model as long as its performance does not degrade.

The next step implements what is known as the $\epsilon$-greedy strategy in
reinforcement learning \citep{sutton1998introduction} and is one of the most
simple strategies to balance exploitation and exploration of an agent. The
strategy executes the action with the highest reward with a probability of
$1-\epsilon$ and a random action with a probability of $\epsilon$.
Additionally, we anneal $\epsilon$ from $1.0$ to $0.1$ during the first couple
of iterations. We chose to anneal $\epsilon = 0.9 e^{-\alpha n}+0.1 \propto
e^{-n}$ where $n$ is the number of iterations and $\alpha$ such that
$\epsilon=0.5$ at $n=5$.

If a greedy step is selected, a grid search over the current model for all
legal positions is performed to locate the global maximum. We chose grid search
because the prediction steps of internal models are usually fast enough in
relation to movement speeds and model learning to allow for brute force. Any
standard optimization method could be used as a drop-in replacement.  The robot
then moves either directly towards the maximum if it is not further away than
the step width or moves one step width towards the maximum. If the step leads to
an invalid position, a random search on the line between the current position
and the maximum starting at one step width distance is performed. The first valid point on
this line that the search finds is chosen, i.e., either directly in front or
after the obstacle blocking the step.

An $\epsilon$-step means a random movement. We chose to implement that random
movement in a novelty-driven way choosing only points where the robot has not
been before. This is done by checking if a possible random movement is a
minimum distance of about the size of the robot away from any points of the
past trajectory of the robot and discarding those options. Points outside the
known map are also discarded because of practical reasons.  A point $\vec{x}$ is
chosen randomly under these constraints from a probability distribution
$P(\vec{x}) \propto e^{-\lVert\vec{x}-\vec{x}_0\rVert}$ where $\vec{x}_0$ is
the current position of the robot.

As for all optimization problems, there is no canonic way to terminate it. The
most straightforward options would be a cut-off signal strength, which can be
problematic because of noise, a maximum number of iterations or manual
stopping, which we used in the experimental examples. The algorithm itself will
stop moving the robot once a global maximum of the model has been reached. The
algorithm continues and will occasionally lead to $\epsilon$-steps which drive
the robot away from the maximum. If the predicted maximum is truly a maximum, the
algorithm will lead the robot back to this maximum after executing the
$\epsilon$-step. If however the new measurements collected by this
$\epsilon$-step contradict the model predictions, the model will be updated and
possibly predict a different maximum. Thus global convergence can be assumed.

In general the parameters of the algorithm are not critically important for
convergence, they mainly balance exploitation against exploration and are thus
to some degree purely design choices. We tested different parameter
configurations without changing the behaviour of the algorithm fundamentally.
We settled on a configuration which worked well with the constraints we had for
computational speed, speed of the robot and the test environment.

\section{Internal Model View}

The concept of internal models originates from control theory but has been
introduced in the fields of biology
\citep{wolpert2011principles,haruno1999multiple} as well as in robotics
\citep{demiris2006hierarchical,Schillaci:2012}.  Internal models are often used
as a representation of sensorimotor skills, and usually consist of a pair of
forward model (predictor), which predicts sensory states as a consequence of
motor commands performed at a current state, and inverse model (controller),
which provides motor commands leading to a desired sensory state.  Here we are
mainly interested in forward models, which we will call internal models for the
sake of simplicity. The inverse models in our case are predefined mappings of
positional information, signal strength and steering commands as defined in
\cref{sec:algorithm}.

A prominent hypothesis in Cognitive Robotics and Neuroscience states that
situations beyond a certain minimal complexity can only be effectively handled
by utilising agent-internal models (internal simulations) \citep{little13}.
Internal simulations can cope with noise and delay that is inherent in most
biological and robotics systems.

It has been proposed that internal simulation processes can be behind the
capability of anticipating and of recognising others' actions
\citep{demiris2006hierarchical,wolpert2003unifying,Schillaci:2012}. In the
context of network robotics, this idea can be mapped to using the internal
model of the signal strength of a network node to identify this node. 

Internal models can also be reused since they encode knowledge of the agent.
Internal models of the signal strength distribution of two nodes learned while
trying to locate both nodes could for example be reused in the task of trying
to optimally bridge these two nodes.

The concept of internal models is agnostic to the actual representation of the
internal model, so representations of varying complexity and statistical power
can be used depending on the available resources and prior knowledge. The
representations of internal models reach from simple k-Nearest Neighbors over
linear models or \ac{MLP} to Gaussian Process models.

\subsection{Comparison to Gradient Based Algorithms}

Gradient-based methods are based on the implicit Taylor expansion of the
underlying signal strength function. Using the gradient means using a local
linear approximation of this function, which is fitted every iteration to the
current local data. This is usually done directly on noisy data, and under some
constraints, convergence for this class of algorithms can be proven
\citep{atanasov2012stochastic,blum2014gradient}. One precondition for
convergence is that local maxima, which can result from small scale fading,
have to be mitigated by using more samples for example by fitting a plane to
the data instead of directly calculating gradients using finite differences,
else the algorithms can get stuck in these maxima.  From the internal model
point of view, even plain gradient-based methods make use of an (implicit)
internal model by this local linear approximation even though no information
is propagated from iteration to iteration.

By making the choice of the (internal) model explicit and using all the
available data, we can directly detect and mitigate local maxima.
This also means that we can choose the shortest possible path to the maximum
instead of following the gradient.  The price for this additional information
is additional computational and memory cost which is why we chose to update the
model only when necessary and not every iteration as a gradient-based method
would. This is also due to being able to explicitly check model predictions
against real measured data.

Additionally, the choice of the $\epsilon$-greedy strategy guarantees that the
robot cannot get stuck in a wrongly perceived local maximum which is mistaken
as a global maximum by the internal model. This strategy is a standard strategy
employed in the field of reinforcement learning to balance exploration against
exploitation \citep{sutton1998introduction}.

\section{Implementation}

\subsection{Hardware and ROS}
\label{sec:hardware_ros}

\begin{figure}[t]
  \centering
  \includegraphics[width=0.5\textwidth]{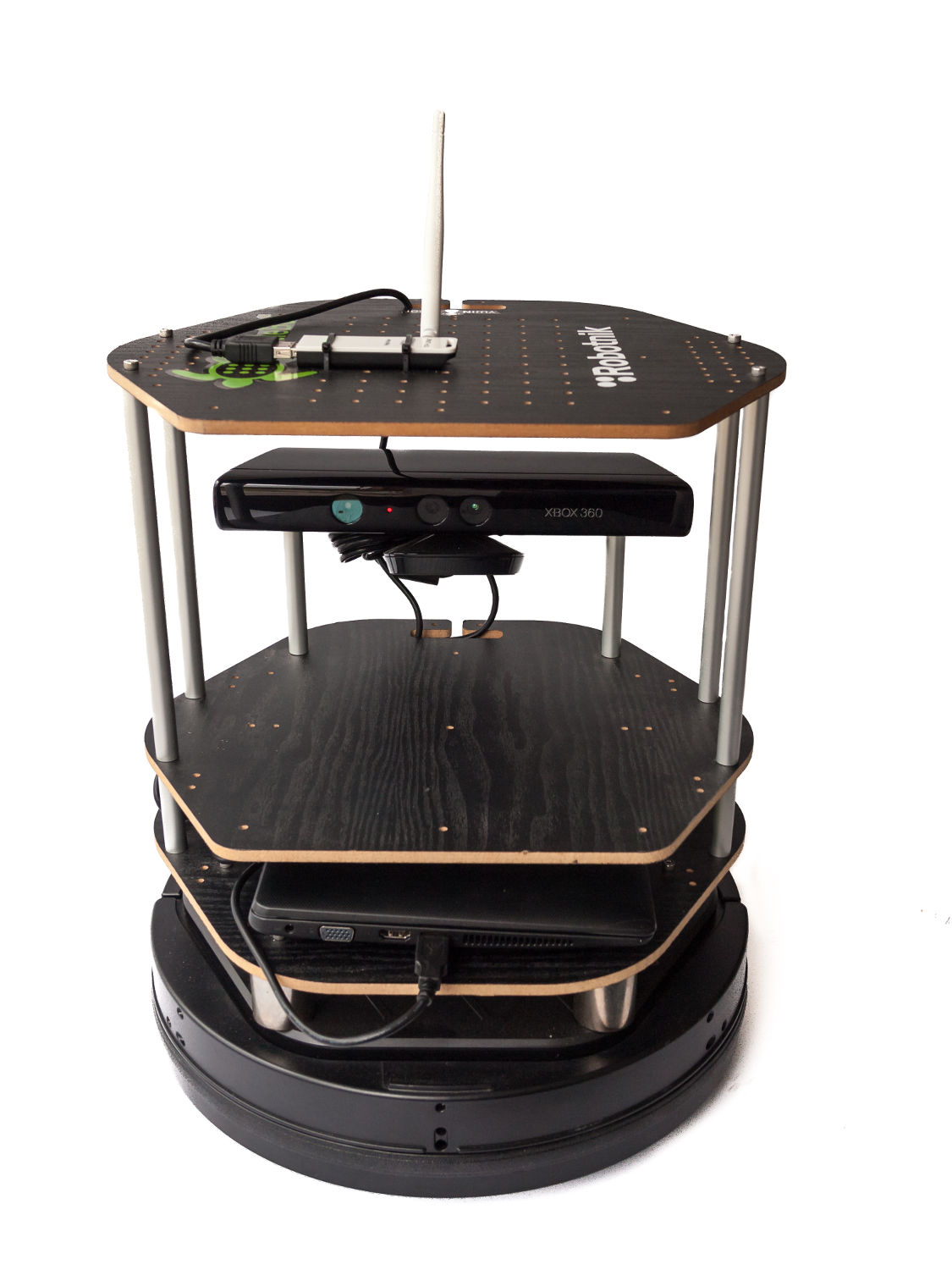}
  \caption{TurtleBot 2 with USB wireless adapter.}
  \label{fig:turtlebot}
\end{figure}

We are using a TurtleBot 2\footnote{http://turtlebot.com/} as a mobile base.
The algorithm is implemented as a ROS node
using a ROS node for capturing and analyzing the packets
We are using the ROS navigation stack\footnote{http://wiki.ros.org/navigation} for
navigation, localization and path planning using the Microsoft Kinect and
odometry. The navigation is facilitated by a pre-learned map using OpenSlam's
Gmapping\footnote{http://wiki.ros.org/gmapping} and
amcl\footnote{http://wiki.ros.org/amcl}.  As a medium we are using
off-the-shelf usb 802.11 wireless dongles and
tcpdump/libpcap\footnote{http://www.tcpdump.org/} for capturing packets. For
all captured packets sent via the correct MAC-address, \ac{RSSI} values are
extracted from the Radiotap headers. The robot is depicted in \cref{fig:turtlebot}.

\subsection{Internal Models}
\label{sec:internal_models}

To limit the computational cost of learning an internal model 
we chose to learn on a maximum of randomly drawn $10,000$ samples from our data.
In the experiments we easily exceed $30,000$ samples. In the following two
subsections, \cref{sec:internal_models:ridge} and
\cref{sec:internal_models:mlp}, we give a short overview over the two chosen
instantiations of internal models and their implementation as learning
strategies.

\subsubsection{Local Ridge Regression}
\label{sec:internal_models:ridge}

Ridge regression is a linear least squares model with a Tikhonov regularization
\citep{tikhonov1943stability}. We use a local version of this by only fitting it
to data inside of some radius to have a similar model as used in gradient-based
algorithms. We used Ridge regression implemented by scikit-learn
\citep{scikit-learn} with a local radius of \SI{5}{\metre} and regularization
parameter $\alpha=1.0$. No preprocessing of the data was used.

Even though using this model is similar to the idea of a locally linear Taylor
approximation as used in gradient-based methods, it is superior in that it
averages over an area, mitigating the effects of small scale fading and noise
and in that it is regularized in order to prevent instabilities and
overfitting.

\subsubsection{Multilayer Perceptron}
\label{sec:internal_models:mlp}

As a second example of an internal model we chose an \ac{MLP} with
two input neurons, a hidden layer of 100 sigmoid neurons and one linear output
neuron. All layers also have access to a bias node. Classic backpropagation
with a small momentum term\footnote{learning rate $0.01$ and momentum $0.1$}
for three epochs was used for learning. We chose to use a fixed number of
epochs to limit computational cost. We are not interested in full convergence
and cross-validation since the algorithm checks the predictions of the
model against real measured data constantly and re-learns the model if the
prediction errors surpass a threshold. The network and learning was
implemented using PyBrain \citep{pybrain2010jmlr}.  We preprocessed the data
to remove the mean and scale it to unit variance because those are assumed by
our implementation of artificial neural networks.

\section{Experiments}

All experiments are performed in a regular office environment with about
$20$ different active access points and countless clients in the \SI{2.4}{\giga\hertz} band
in the test environment. The source node was emitting around
$200$ dummy $\text{packets}/s$ which were then captured by the robot.

\begin{figure}[t]
  \centering
  \includegraphics[width=0.5\textwidth]{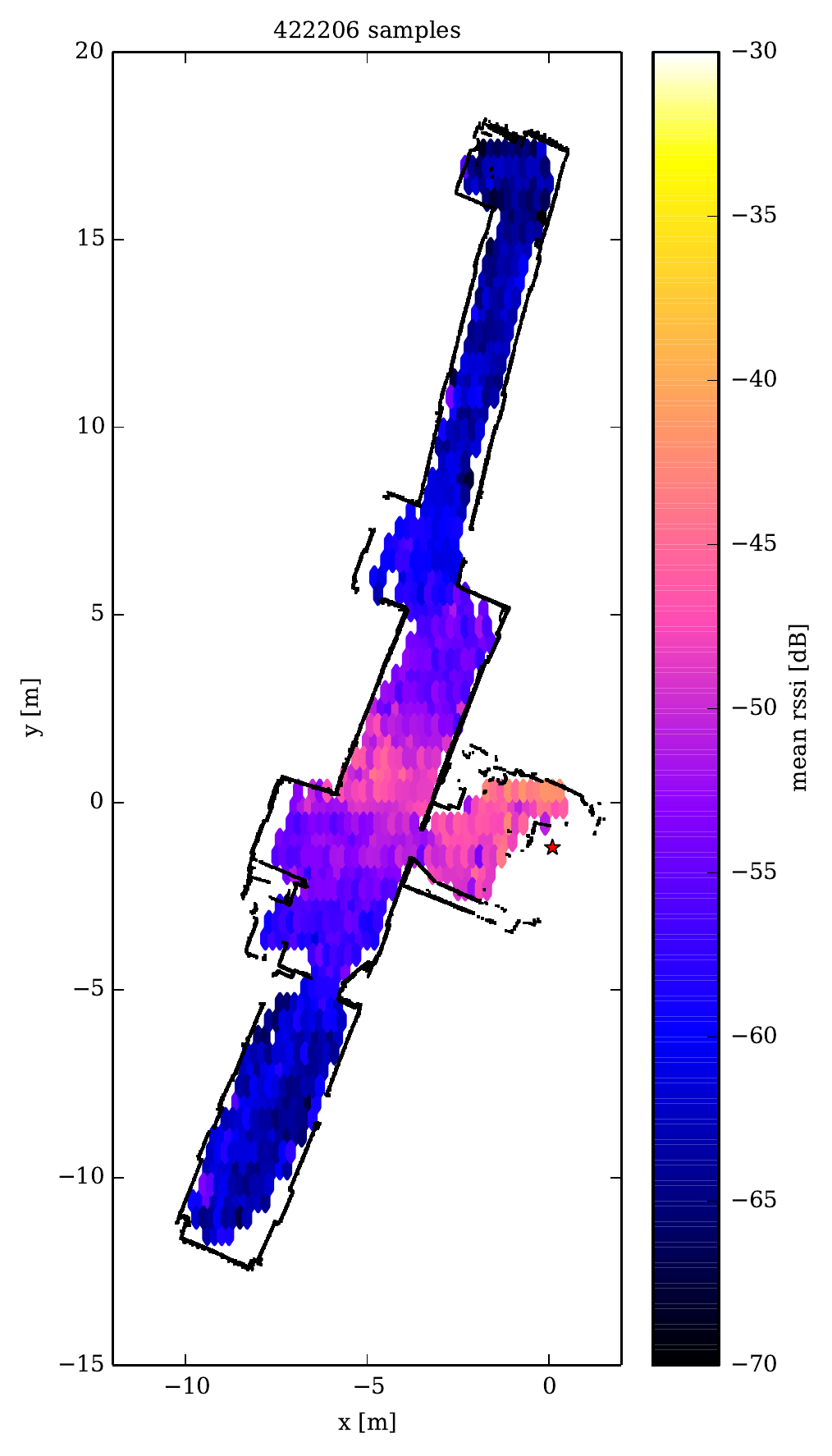}
  \caption{Histogram of the example target function $f$ with $422,206$
  measurement samples. The position of the target node is depicted as a red
  star.}
  \label{fig:rssi_hexbin}
\end{figure}

We also did a complete mapping of the experimental area in terms of \ac{RSSI}
using our setup. The resulting hexagonal histogram is depicted in
\cref{fig:rssi_hexbin}. The bins of the histogram are larger than the
wavelength of the wireless signal which means that it represents pure path loss
and no interference effects such as small scale fading.  We measured an
average logarithmic Gaussian noise with a standard deviation of \SI{4}{\deci\bel}.

To accurately model small scale fading and other interference effects, a very
accurate simulation or even the complete solution of Maxwell's equations is
necessary. This is computationally costly and requires accurate knowledge of
material properties of walls etc. Effective models on the other hand are cheaper
to calculate but only represent an abstract, stochastic model of the
environment. Thus we chose to work directly with real world experiments in order not
to lose or miss systematic physical effects such as small scale fading.

Since we performed the experiments during regular office hours, a lot of
disturbances such a persons walking, opening and closing doors, changing
network load etc. affected the measurements and path planning. Additionally,
the algorithm itself is stochastic. Thus, a large number of experiments would
have to be conducted to gain statistically sound results. We chose to show some
typical experiments instead.

\begin{figure}[t]
  \begin{center}
    \hspace*{\fill}
    \includegraphics[width=0.19\textwidth]{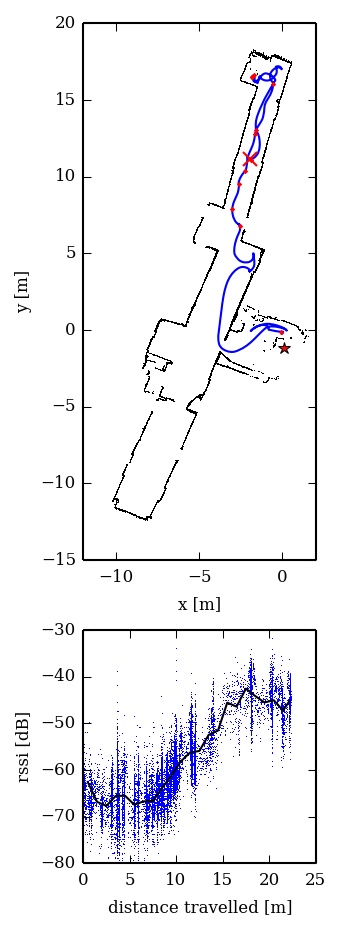}\hfill
    \includegraphics[width=0.19\textwidth]{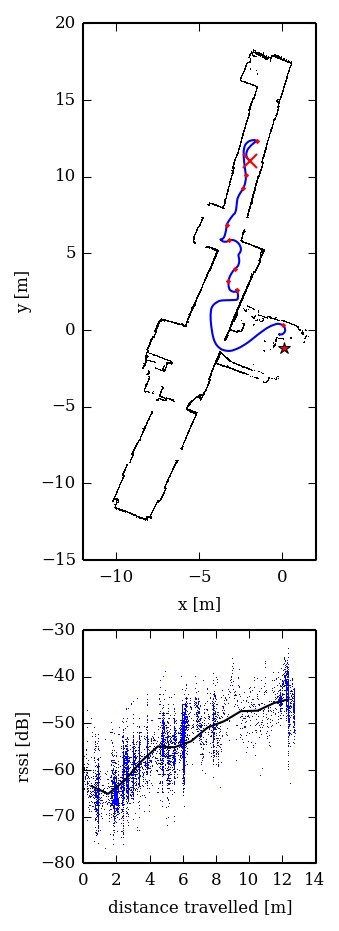}\hfill
    \includegraphics[width=0.19\textwidth]{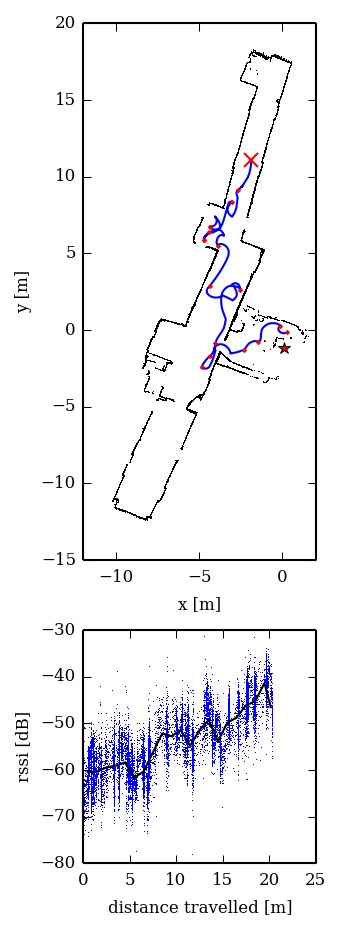}\hfill
    \includegraphics[width=0.19\textwidth]{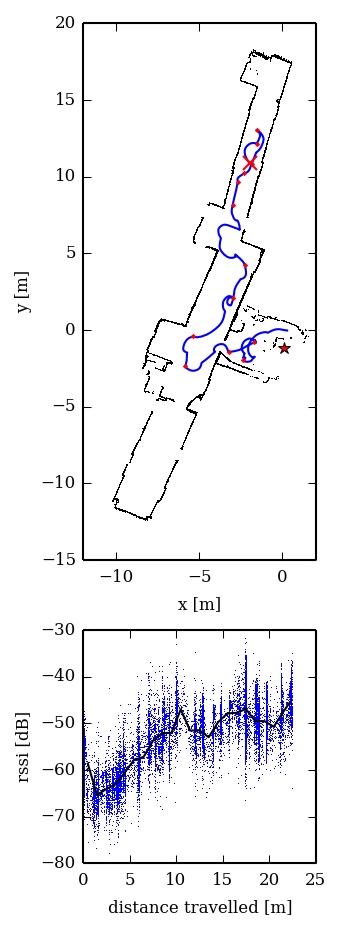}\hfill
    \includegraphics[width=0.19\textwidth]{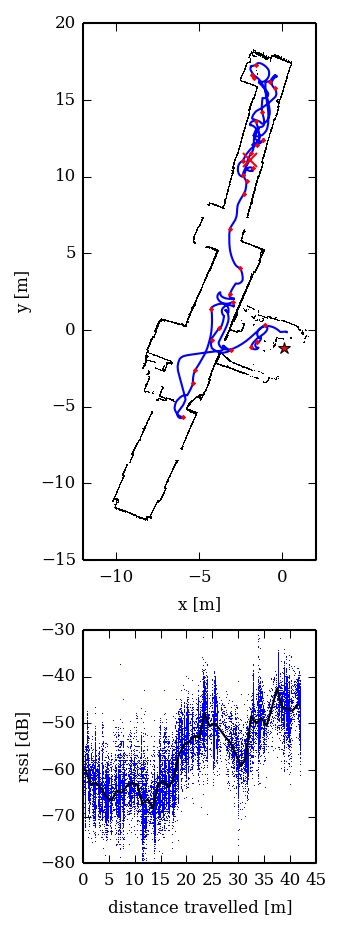}
    \hspace*{\fill}
  \end{center}
  \caption{Experiments done with Ridge regression as an internal model. Top
  panels shows trajectories on the map of the used office space and lower panels
  show the history of the measured \ac{RSSI} values as a function of the
  distance covered and the moving average over \SI{1}{\metre}. The red star denotes the
  position of the source. Red circles on the trajectory denote model updates.
  The initial position of the robot is marked with a red cross.}
  \label{fig:exp_ridge}
\end{figure}

\begin{figure}[t]
  \begin{center}
    \hspace*{\fill}
    \includegraphics[width=0.19\textwidth]{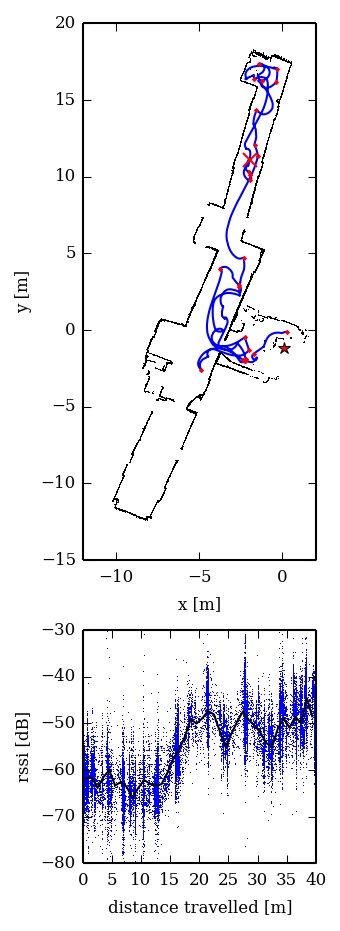}\hfill
    \includegraphics[width=0.19\textwidth]{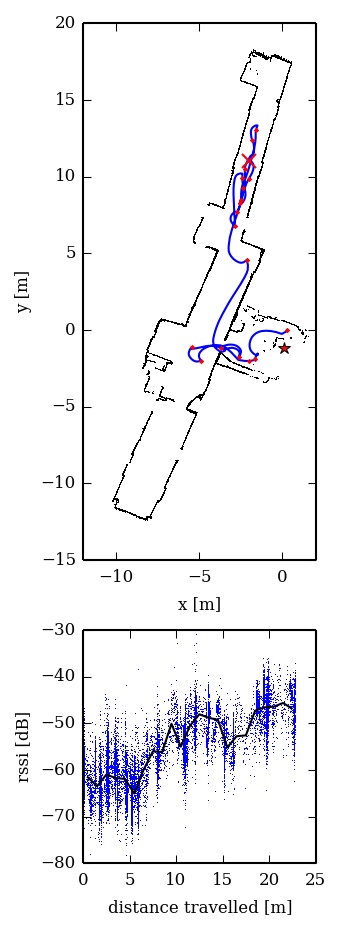}\hfill
    \includegraphics[width=0.19\textwidth]{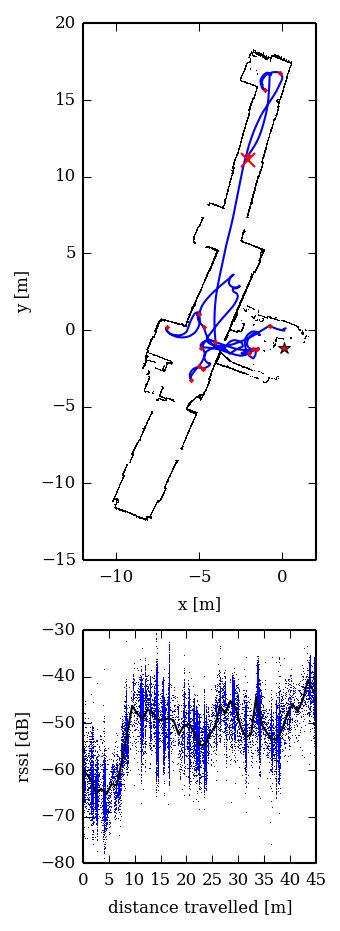}\hfill
    \includegraphics[width=0.19\textwidth]{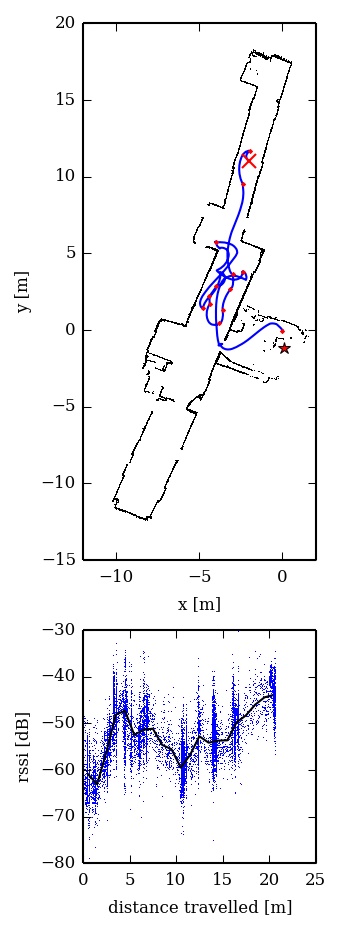}\hfill
    \includegraphics[width=0.19\textwidth]{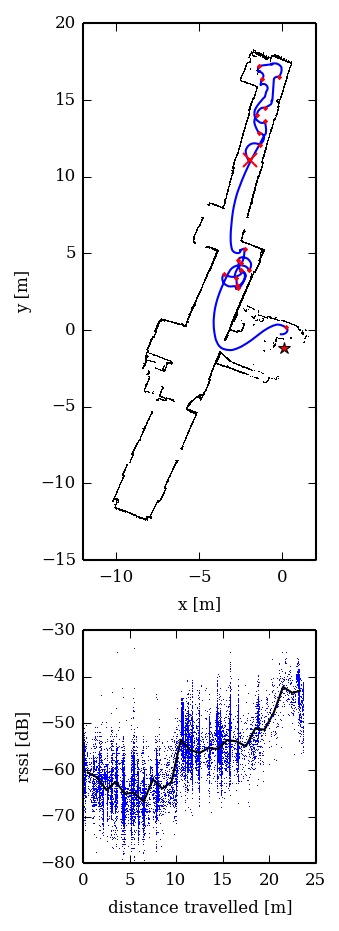}
    \hspace*{\fill}
  \end{center}
  \caption{Experiments done with an \ac{MLP} as an internal model.  Top panels
  shows trajectories on the map of the used office space and lower panels show
  the history of the measured \ac{RSSI} values as a function of the distance
  covered and the moving average over \SI{1}{\metre}. The red star denotes the position
  of the source. Red circles on the trajectory denote model updates.  The
  initial position of the robot is marked with a red cross.}
  \label{fig:exp_fnn}
\end{figure}

Five typical runs for the two different internal model implementations
discussed in \cref{sec:internal_models} are depicted in
\cref{fig:exp_ridge} and \cref{fig:exp_fnn} respectively. 
For all experiments the robot starts from the same initial position.  The plots
show trajectories, points where the model was updated as well as the signal
strength as a function of the way travelled\footnote{Duration times vary wildly
because of the robot stopping for persons, different training times, etc.}.

The trajectories show several distinct variations which can shed some light on
how the algorithm works. In the beginning of each experiment the robot does not
posses any knowledge about the signal strength distribution and starts with an
initial random step. For the next five iterations $\epsilon$ is higher than
$0.5$ and since the $\epsilon$-steps are novelty driven, the probability of
continuing in the same direction as the initial random step is higher than the
probability of turning around. Once $\epsilon$ is approaching its final value
of $0.1$ and the robot has collected enough samples of the signal strength
distribution, it either turns around if it initially went into the wrong
direction or it continues on towards the maximum. Once the robot passes the
room where the target node is located, it shows a similar behaviour. Either the
internal model correctly predicts the position of the target node and the robot
directly enters the room or the robot passes by the room but then turns around
after the prediction error of the internal model measured against the collected
samples passes the error threshold and the model is updated with new samples.

Considering the measurement history of the experiments, it is clear how big of
a challenge the task is. Especially far away from the source, the gradients are
much smaller than the noise of the measurements. Using finite difference
gradient methods would only work for impractically large measurement steps or
would lead to very noisy trajectories close to biased random walks. In
contrast, both internal models seem to be able to cope with the noise and
possible local maxima.

The particular choice of the internal model does not seem to make much of a
difference though. This may be due to the simple (and convex) nature of the
underlying function, which is essentially the path loss function measured and
depicted in \cref{fig:rssi_hexbin}. The choice of a concrete model might be
more important when working with more complex target functions as discussed in
\cref{sec:extensions}.  In addition to the two exemplary models shown here,
we also tried Support Vector Regression, Kernel Ridge Regression, different
other configurations of an \ac{MLP}, k-Nearest Neighbors Regression and Radius
Nearest Neighbor Regression. All representations converged albeit with
different convergence speeds. We believe that the choice of the meta-algorithm,
i.e., constant model validation and novelty driven $\epsilon$-greedy search,
will lead to convergence of the algorithm as long as the particular model is in
principle able to represent the target function.

\section{Extensions}
\label{sec:extensions}

\subsection{Idea und Theory}

\begin{figure}[t]
  \centering
  \includegraphics[width=0.8\textwidth]{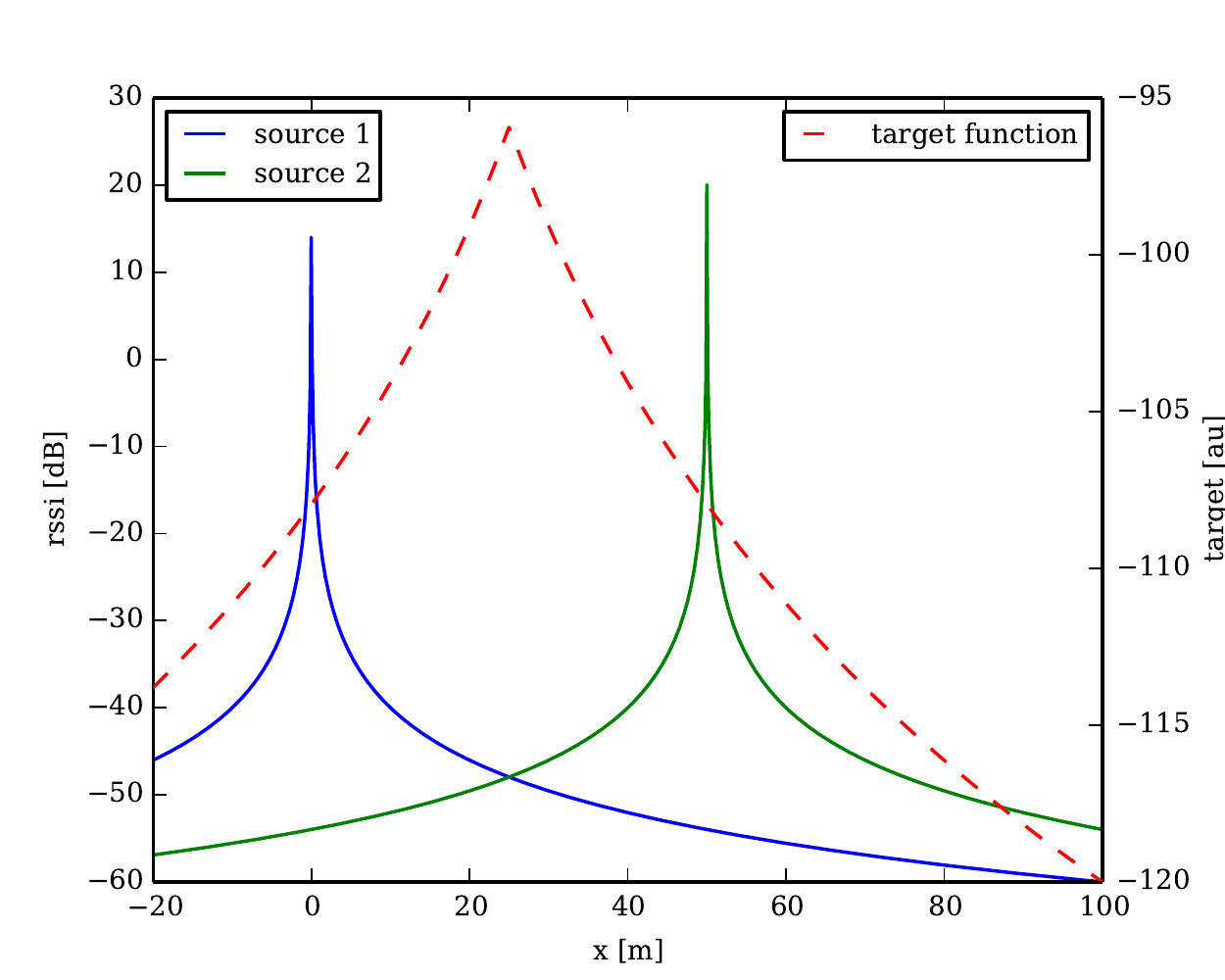}
  \caption{Example target function $f$ for finding the optimal position to
  bridge two network nodes as a 1-dimensional simplification. The signal
  strength of the two nodes with a toy path loss models are plotted on the left
  y-axis. The target function $f$ is plotted on the right y-axis.}
  \label{fig:extension_example}
\end{figure}

The presented algorithm can readily be extended to solve different tasks by
noting that in essence it is just maximizing an unknown function with access to
point measurements. The most straightforward extensions in the networking
context would thus be to replace signal strength with any other interesting
metric like for example \ac{PDR} or \ac{BER}. These metrics might be more noisy
and contain a lot of local minima but the algorithm is able to cope with that
by design.

More complex tasks can be easily constructed using compound metrics. As an example
we consider the task of bridging two network nodes. Practically, this means
moving to the point with the maximal but equal signal strength to both nodes.
We construct a single metric fulfilling this criteria using the signal
strengths of both nodes using a target function $f$ like

\begin{equation*}
  f = -\lvert\text{RSSI}_1 - \text{RSSI}_2\rvert - \lvert\text{RSSI}_1 + \text{RSSI}_2\rvert
\end{equation*}

where the indices $1$ and $2$ correspond to both nodes in question. This target
function $f$ with toy signal strength path loss functions for both nodes is
depicted in \cref{fig:extension_example}.

The target function shows a clear maximum at exactly the center between the two
source positions, so maximising it leads to maximal and equal signal strengths
of both nodes.

This is only a toy example and other task-specific metrics can be designed
making this algorithm very versatile in formulating and solving tasks for
mobile nodes in wireless networks. A cross-layer approach would be especially
interesting. For more complicated target functions, which are probably also
more noisy, more complex internal models can be beneficial.

\subsection{Experiment}

\begin{figure}[t!]
  \centering
  \includegraphics[width=0.5\textwidth]{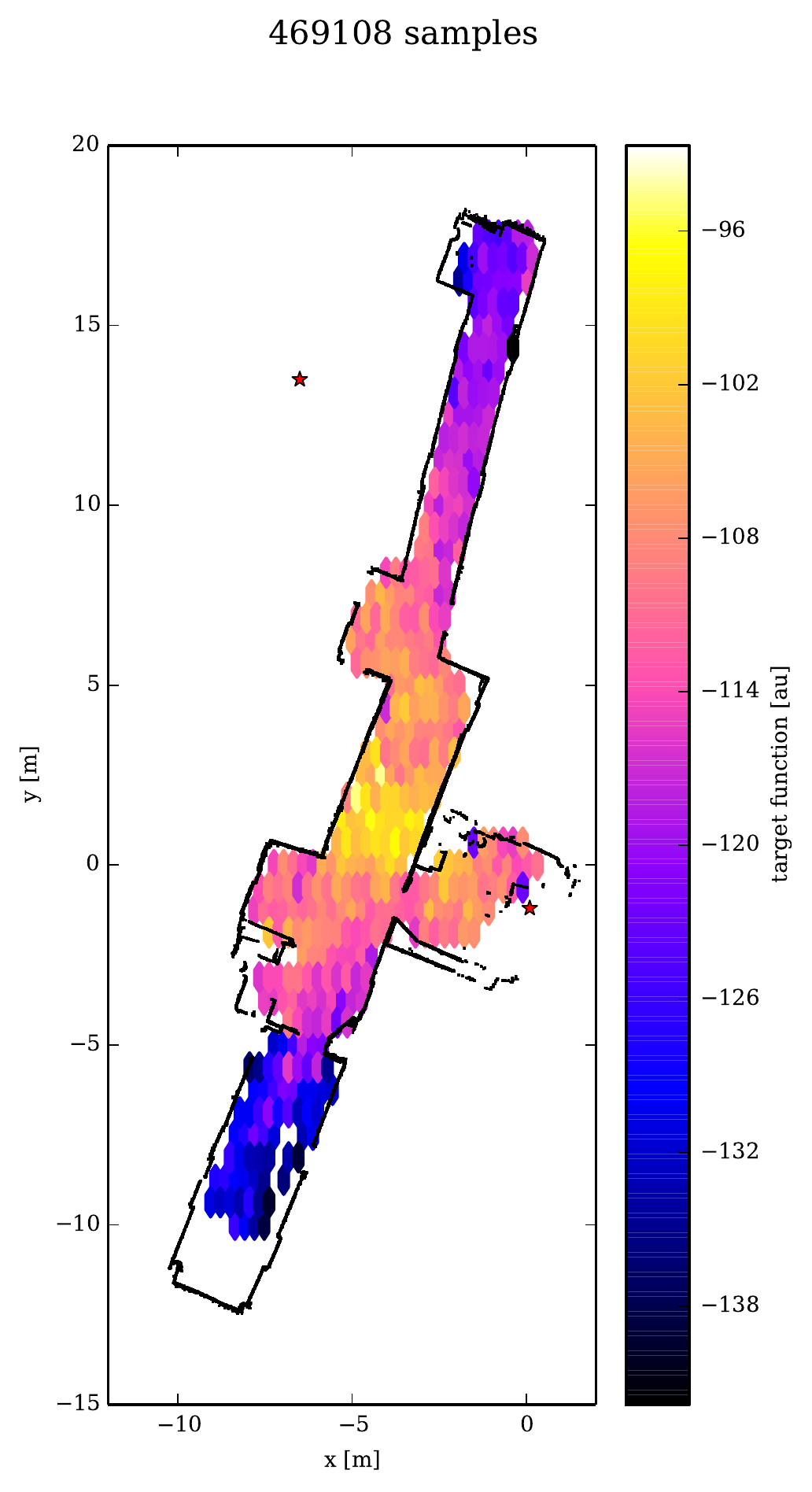}
  \caption{Histogram of the example target function $f$ with $469,108$
  measurement samples. The positions of the two nodes to be bridged are
  depicted as red stars.}
  \label{fig:extension_hexbin}
\end{figure}

\begin{figure}[t!]
  \centering
  \includegraphics[width=0.8\textwidth]{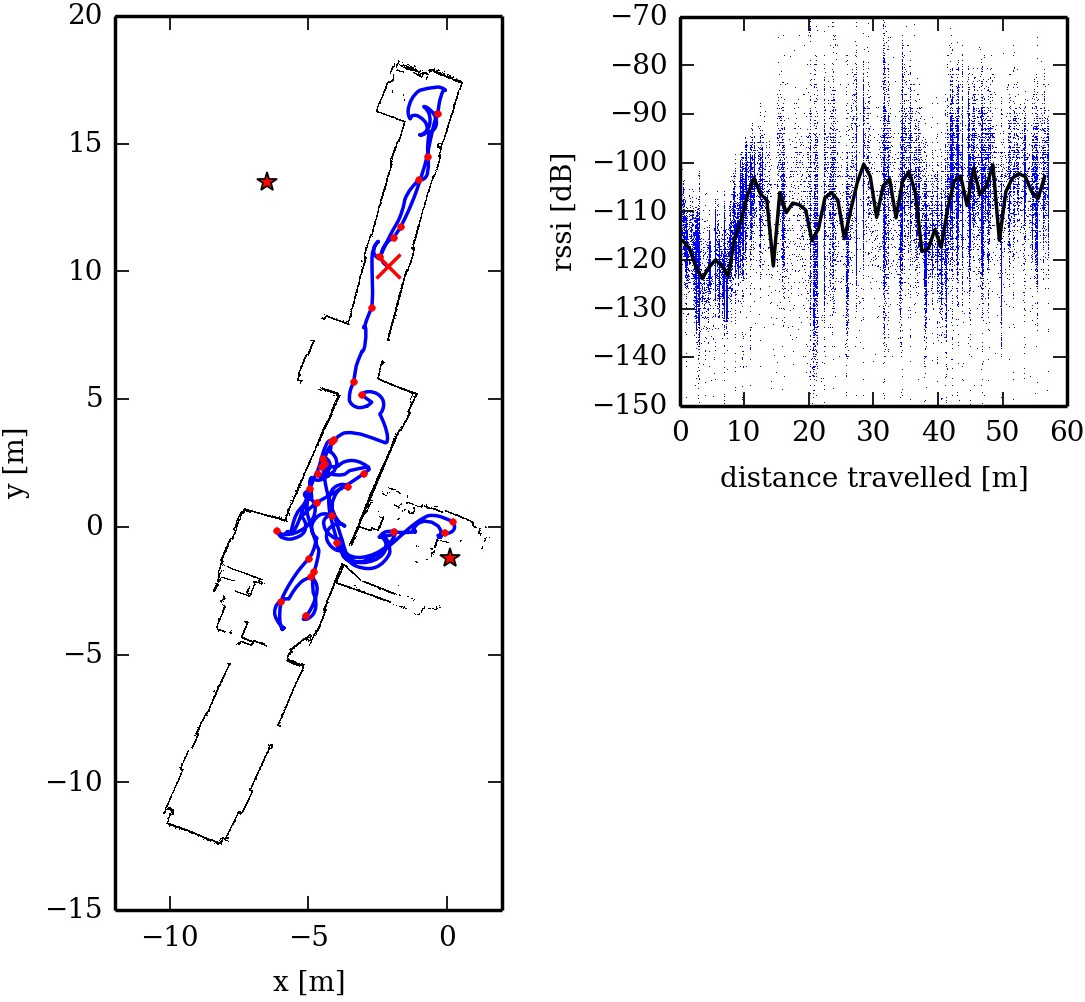}
  \caption{The top panel depicts the trajectory of the robot for one experiment
  using the target function $f$. The initial position of the robot is marked
  with a red cross.  The positions of the two nodes which have to be bridged
  are depicted as red stars. Model updated are denoted by red points. The lower
  panel shows the measurements of the target function as a function of
  travelled distance and the moving average over \SI{1}{\metre} of the measurements.}
  \label{fig:extension_experiment}
\end{figure}

We did exemplary experiments with two nodes and the target function shown
above. One of the nodes was the one used for the earlier experiments and the
second one was placed in a different office. Both nodes were simple USB wireless adapters
dongles sending packets. Again, the experiments were conducted during regular
office hours. \cref{fig:extension_hexbin} shows the histogram of the
measured values over all experiments conducted. As expected the function has one
maximum somewhere between both nodes.

Using the target function $f$ means that the noise of the measurements of both
sources are combined following the law of propagation of uncertainty, which
yields for the case of the same but independent noise for both nodes and the
use of the target function $f$ twice the noise of a single node, which is
around \SI{4}{\deci\bel}.  We measured a standard deviation of about \SI{9}{\au} for our
experiments which agrees well with the predicted error.

This also means that the magnitude of measured gradients would be less than the
noise of the experiment, which means that a gradient descent algorithm would
show a behaviour similar to a random walk with drift instead of a noisy
gradient descent. Convergence would not be impacted by this high error but
convergence speed would be low.

\cref{fig:extension_experiment} depicts an exemplary experiment. We used an
\ac{MLP} as the internal model with the same parameters as for the single source
experiment but we changed the error threshold to \SI{7}{\au} to account for the
increased noise. The trajectory of the robot converges to the area of the
maximum of \cref{fig:extension_hexbin}. The experiment was terminated before
a stable maximum was reached because we chose to anneal $\epsilon$ to $0.1$
instead of $0.0$ which means that no fixed position can be reached by design.
The magnitude of the noise can be seen in the measurement log of
\cref{fig:extension_experiment}. 

The behaviour of the algorithm certainly can be improved by tuning parameters
but the general idea of testing the internal model predictions against real
measurements and refining it when necessary in combination with the
$\epsilon$-greedy strategy ensures convergence even without optimized
parameters and in spite of very noisy measurements.

\section{Conclusions}

In this paper we have introduced a class of algorithms to solve a number of
tasks related to network robotics, specifically to autonomous mobile network nodes
interacting with a wireless network. Beginning with the most basic task of
source seeking, the general algorithm has shown to be effective in real world
scenarios by a series of experiments with a real physical robot in an office
environment. We then showed how this algorithm can be extended to various other
tasks, which can be encountered by a mobile network node, such as maximizing
the \ac{PDR} to a certain target node or bridging two network nodes. For the
bridging scenario an exemplary experiment has been conducted successfully.

Furthermore, we have given insight into the robotics view on the problem and
have discussed the relationship to conventional algorithms. We believe that
this discussion can further the field of network robotics in an
interdisciplinary manner and can also lead to new insights in other related
fields.

\section*{Acknowledgments}
We thank everybody who contributed to the success of this project. This
includes all members of the Adaptive Systems Group and the DFG graduate
research training group METRIK (GRK 1324), which also funds one of the authors. 

\bibliographystyle{apalike}
\bibliography{paper}

\begin{acronym}
  \acro{PDR}{Packet-Delivery Ratio}
  \acro{BER}{Bit-Error Rate}
  \acro{RSSI}{Received Signal Strength Indicator}
  \acro{MLP}{Multilayer Perceptron}
\end{acronym}

\end{document}